\documentclass[runningheads]{llncs}
\usepackage{subfigure}
\usepackage{bm}
\usepackage{amsmath}
\usepackage{amssymb}
\usepackage{mathrsfs}
\usepackage{graphicx}
\usepackage{wrapfig}
\usepackage{color}
\usepackage{booktabs}
\usepackage{url}
\usepackage{multirow}
\definecolor{Todocolor}{RGB}{125,125,0}

\usepackage[letterpaper=true,colorlinks,bookmarks=false]{hyperref}
\begin{document}
\title{SkrGAN: Sketching-rendering Unconditional Generative
Adversarial Networks for Medical Image
Synthesis}
%
%
\author{Tianyang Zhang\inst{1,2} \and
Huazhu Fu\inst{3}* \and
Yitian Zhao\inst{2}* \and
Jun Cheng\inst{2,4} \and
Mengjie Guo \inst{5}\and \\
Zaiwang Gu\inst{6}\and
Bing Yang \inst{2}\and
Yuting Xiao\inst{5} \and
Shenghua Gao \inst{5} \and
Jiang Liu\inst{6}
}

\institute{
University of Chinese Academy of Sciences,\and
Cixi Instuitue of Biomedical Engineering, Chinese Academy of Sciences, \email{yitian.zhao@nimte.ac.cn}\and
Inception Institute of Artificial Intelligence, \email{hzfu@ieee.org}\and
UBTech Research\and
ShanghaiTech University\and
Southern University of Science and Technology
}
\maketitle              
\begin{abstract}
	
Generative Adversarial Networks (GANs) have the capability of synthesizing images, which have been successfully applied to medical image synthesis tasks.  However, most of existing methods merely consider the global contextual information and ignore the fine foreground structures, e.g., vessel, skeleton, which may contain diagnostic indicators for medical image analysis. Inspired by human painting procedure, which is composed of stroking and color rendering steps, we propose a Sketching-rendering Unconditional Generative Adversarial Network (SkrGAN) to introduce a sketch prior constraint to guide the medical image generation. In our SkrGAN, a sketch guidance module is utilized to generate a high quality structural sketch from random noise, then a color render mapping is used to embed the sketch-based representations and resemble the background appearances. Experimental results show that the proposed SkrGAN achieves the state-of-the-art results in synthesizing images for various image modalities, including retinal color fundus, X-Ray, Computed Tomography (CT) and Magnetic Resonance Imaging (MRI). In addition, we also show that the performances of  medical image segmentation method has been improved by using our synthesized images as data augmentation.

\end{abstract}

\section{Introduction}

In the last decade, deep learning techniques  have shown to be very promising in many visual recognition tasks \cite{fu2018joint,gu2019net}, including object detection, image classification, face recognition, and medical image analysis. The large scale training data is extremely important for training accurate and deep models. Although it is easy to collect data in conventional computer vision tasks, it is often difficult to obtain sufficient high quality data in medical imaging area.
Recently, Generative Adversarial Networks (GANs) are proposed to generate a distribution that matches the real data distribution via an adversarial process~\cite{Goodfellow2014Generative}. Due to the powerful capability of image generation, GANs have been successfully applied to many medical image synthesis tasks, including retinal fundus~\cite{Costa2017End,He2018Synthesizing}, X-Ray~\cite{Madani2018}, CT and MRI images~\cite{zhang2018translating} synthesizing.

 \begin{figure}[!t]
	\center
	\includegraphics[width=1\textwidth]{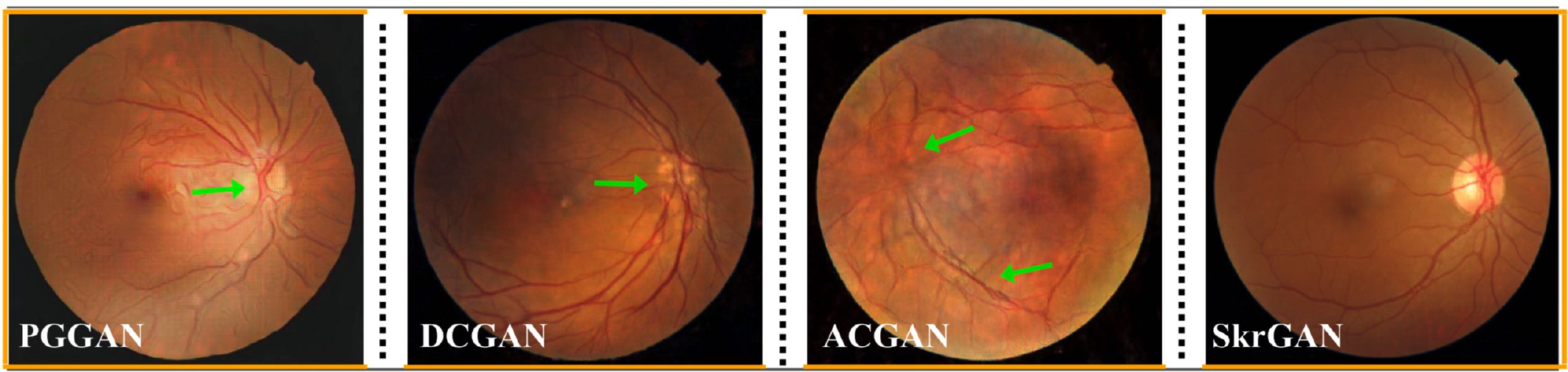}
	\caption{Synthesized retinal images by PGGAN~\cite{PGGAN}, DCGAN~\cite{DCGAN}, ACGAN~\cite{odena2017conditional} and our SkrGAN. Compared with these methods, our method performs better in retaining structural details, e.g., blood vessels, disc and cup regions, as indicated by green arrows.}
	\label{img_cover}
\end{figure}

The GANs algorithms can be divided into the conditional and unconditional manners. The conditional GANs direct the data generation process by conditioning the model on additional information~\cite{mirza2014conditional}, which have been widely used in cross-modality synthesis and conditioned segmentation. For example, the pix2pix method is proposed to translate images from one type to another~\cite{isola2017image}. An auxiliary classifier GAN (ACGAN) is provided to produce higher quality sample by adding more structures to the GAN latent space~\cite{odena2017conditional}.
In~\cite{zhang2018translating}, a CT and MRI translation network is provided to segment multimodal medical volumes. By contrast, the unconditional GANs synthesize images from random noise without any conditional constraint, which are mainly used to generate images. 
For example, Deep Convolutional GAN (DCGAN)~\cite{DCGAN} uses deep convolution structure to generate images.
$\text{S}^2$-GAN \cite{wang2016generative} materializes a two-stage network and depth maps to generate images with realistic surface normal map (i.e, generate RGBD images). However, the $\text{S}^2$-GAN requires depth maps of the training dataset, while we usually do not have medical image datasets with paired depth maps. 
 Wasserstein GAN (WGAN)~\cite{WGAN} improves the loss and training stability of previous GANs to obtain a better performance.
 Progressive Growing GAN (PGGAN)~\cite{PGGAN} grows the depth of convolution layers to produce the high resolution natural images.

In this paper, we aim to generate high quality medical images with correct anatomical objects and realistic foreground structures. Inspired by realistic drawing procedures of human painting~\cite{ostrofsky2012perceptual}, which is composed of stroking  and color rendering, we propose a novel unconditional GAN named  Sketching-rendering Unconditional Generative Adversarial Network (SkrGAN) for medical image synthesis. Our SkrGAN decomposes into two tractable sub-modules: one sketch guidance module generating the structural sketch from random noise; and one color render mapping module producing the structure-preserved medical images.  The main contributions of this paper are summarized as follows:\\
\textbf{1)} An unconditional GAN, named SkrGAN, is proposed for medical image synthesis. By decomposing the whole image generation into sketch guidance and color rendering stages, our SkrGAN could embed the sketch structural representations to guide the high quality medical image generation. \\	
\textbf{2)} The experiments in four medical imaging modalities synthesizing tasks show that our SkrGAN is more accurate and robust to variations in the size, intensity inhomogeneity and modality of the data than other state-of-the-art GAN methods. \\
\textbf{3)} The medical image segmentation experiments demonstrate that our SkrGAN could be applied as a data augmentation method to improve the segmentation performance effectively.

\section{Proposed Method}

\begin{figure}[!t]
	\includegraphics[width=\textwidth]{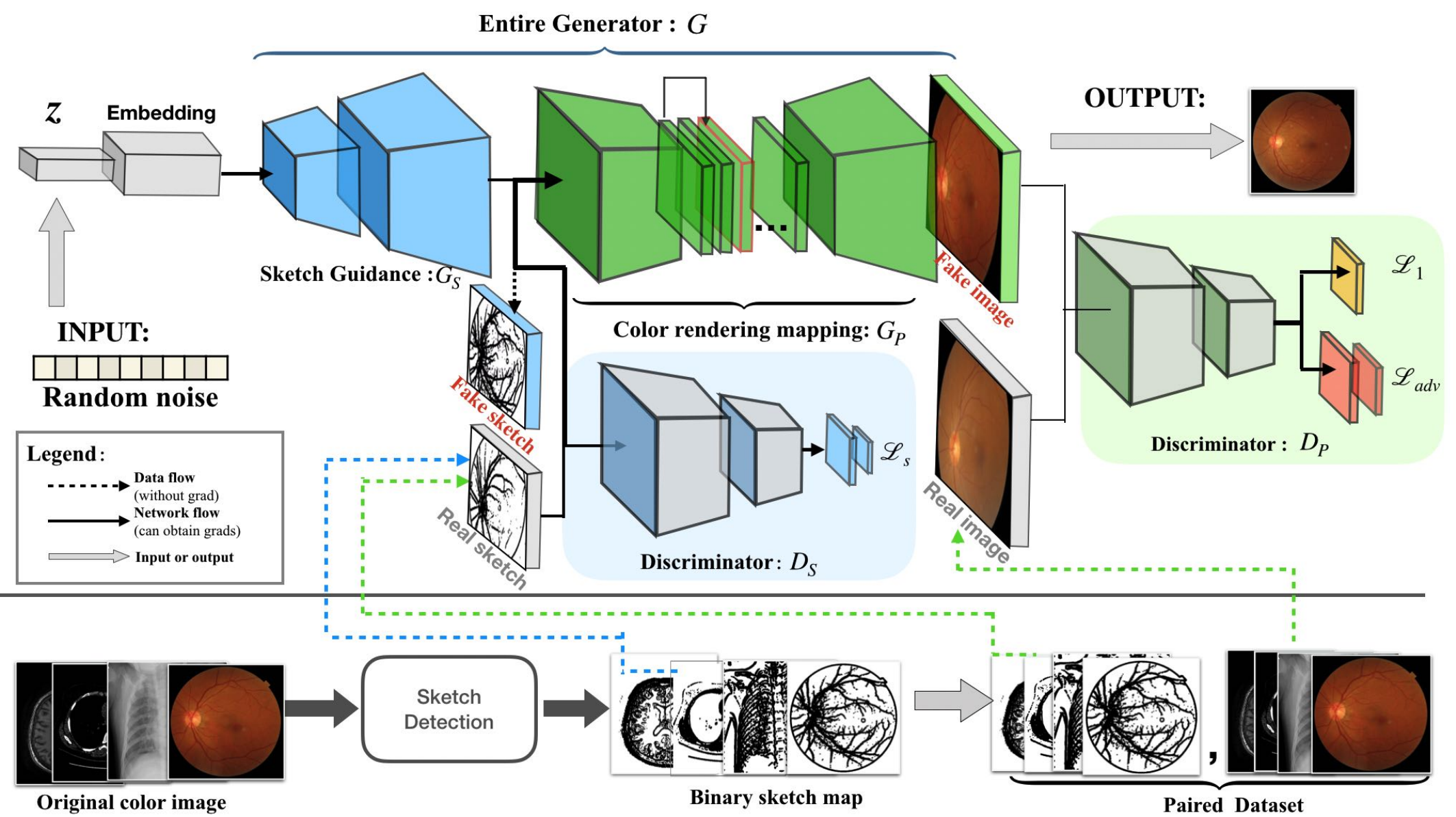}
	\caption{Illustration of our SkrGAN structure, which can generate medical images from the input noises. The sketch guidance module $G_S$ (blue block) obtains the representations based on sketch structure discriminator $D_S$. The color render mapping $G_P$ (green block) embeds the sketch representations to generate the final color image with a discriminator $D_P$. Moreover, We also extract a sketch draft dataset (bottom) for training the model. (best viewed in color)}
	\label{img_frame}
\end{figure}

Inspired by realistic drawing skills of the human painting~\cite{ostrofsky2012perceptual}, which suggests that the painting is usually accomplished from simple to difficult procedures, i.e., from sketching to color rendering, we propose a novel Sketching-rendering Unconditional Generative Adversarial Networks (SkrGAN), to generate high quality medical images with realistic anatomical structures. As shown in Fig.~\ref{img_frame}, we decompose the entire image generator $G$ into two phases, as a sketch guidance module $G_S$ (in Sec.~\ref{sec_sketch}) and a color render mapping $G_P$ (in Sec.~\ref{sec_color}) . The sketch guidance module $G_S$ generates the sketch structural representations with a sketch discriminator $D_S$, while the color render mapping $G_P$ embeds the sketch representations to generate the final image with a color discriminator $D_P$.

\subsection{Sketch Draft Preparation}
\label{sec_data}

In order to train our SkrGAN, the sketch draft corresponding to the input training image is required by sketch discriminator $D_S$. We aim to retain the main structural information of the given images, such as the blood vessels of retinal fundus, and bones of X-ray images. 
In our method, firstly the Sobel edge detection method is used to extract the initial structural boundaries, and then a Gaussian lowpass filtering is applied to remove the isolated noise and  pixels. Finally, a morphological operation consisting of an opening process followed by a closing process is employed to remove noise further and fill the vessel-like structures. This procedure will greatly reduce the complexity of sketch images, which makes the sketch synthetic process easier than just using traditional edge detection methods. An example of sketch draft detection method could be found at the bottom of Fig.~\ref{img_frame}, where the main sketch structures (e.g., vessels and bones) are extracted.

\subsection{Sketch Guidance Module} 
\label{sec_sketch}

With the given dataset $X$ and corresponding sketch draft set $Y$ by the sketch draft extraction, the sketch guidance module $G_S$ is trained by using loss $\mathcal{L}_s$ in sketch discriminator $D_S$:
\begin{align}
	\begin{cases}
		\mathcal{L}_{s} = 
		 \mathbb{E} _{\bm z\sim \bm p_{noise}}[\log (D_S( G_S (\bm z\odot \bm l )))]+\mathbb{E}_{\bm x\sim p_{\bm x}}[\log(1-D_S(\bm y))] \\
	D_S=D_S^{(n)}\cdots D_S^{(1)}D_S^{(0)}\\
	G_S=G_S^{(n)}\cdots G_S^{(1)}G_S^{(0)}
	\label{equ2}
\end{cases}
\end{align}
where $\bm z\sim\bm p_{noise}$ and $\bm l$  represent the noise pattern and latent code respectively; $\bm p_{\bm x}$ represents the distribution of $\bm x$ and $\odot$ is the element-wise multiplication. $D_S^{(i)}, i=0,1,...,n$ denote discriminating layers of the discriminator in different levels, whose inputs are determined to different resolutions. $G_S^{(i)}, i=0,1,...,n$ are the generating layers of different resolutions, respectively.
More concretely, our method iteratively adds convolutional layers of the generator and the discriminator during the training period, which guarantees to  synthesize images at $\{2^{k+1}\times 2^{k+1}| k= 1,2,...,8\}$ resolutions.  Additionally, the training process fades in the high resolution layer smoothly by using skip connections and the smooth coefficients. For simplicity, we utilize the network structure in PGGAN~\cite{PGGAN} as the backbone of $G_S$.

\subsection{Color Render Mapping} 
\label{sec_color}

The color render mapping $G_P$  translates the generated sketch representations to color images, which contains the U-net~\cite{ronneberger2015unet} structure as backbone, and a color discriminator $D_P$ for adversarial training. Two losses $\mathcal{L}_{adv}$ and $\mathcal{L}_1$ for training $G_P$ are described as:
\begin{align}
	\begin{cases}
		\mathcal{L}_{adv}=\mathbb{E}_{\bm y \sim Y}[\log (D_P(G_P(\bm y),\bm y))]
	+\mathbb{E}_{(\bm x,\bm y)\sim (X,Y)}[\log (1-D_P(\bm x,\bm y))]\\
	\mathcal{L}_1=\lambda \mathbb{E}_{(\bm x,\bm y)\sim (X,Y)}\|G_P(\bm y)-\bm x\|_1\label{eq4}
	\end{cases}
\end{align}
where $(\bm x,\bm y)\sim(X,Y)$ represent the training pair of real image and sketch. The $\mathcal{L}_{adv}$ is utilized to provide adversarial loss for training $G_P$, while $\mathcal{L}_1$ is utilized to calculate the $L1$ norm for accelerating training.
 Finally, the full objective of our SkrGAN is given by the combination of the loss functions in Eq (\ref{equ2}) and Eq (\ref{eq4}):
\begin{align}
	G_S^{*},G_P^{*},D_S^{*},D_P^{*}=\arg[ \underbrace{\min_{G_S} \max_{D_S}~\mathcal{L}_{s}}_{sketch~guidance}+\underbrace{\min_{G_P} \max_{D_P}~(\mathcal{L}_{adv}+\mathcal{L}_1)}_{color~rendering}].
	\label{equ6}
\end{align}

\section{Experiments}

\subsubsection{Datasets:} Three public datasets and one in-house dataset are utilized in our experiments: Chest X-Ray dataset \cite{Kermany2018Identifying}\footnote{\url{https://www.kaggle.com/paultimothymooney/chest-xray-pneumonia}} with 5,863 images categorized into Pneumonia and normal; Kaggle Lung dataset\footnote{\url{https://www.kaggle.com/kmader/finding-lungs-in-ct-data/data/}} with 267 CT images; Brain MRI dataset\footnote{\url{http://neuromorphometrics.com}} with 147 selected images and a local retinal color fundus dataset (RCF) with 6,432 retinal images collected from local hospitals. In our unconditional experiment, we do not need labeling information. 
\subsubsection{Evaluation Metrics:} In this work, we employ the following three metrics to evaluate the performance in the synthetic medical images, including multi-scale structural similarity (\textbf{MS-SSIM}), Sliced Wasserstein Distance (\textbf{SWD}) \cite{SWD}, and Freshet Inception Distance (\textbf{FID}) \cite{PGGAN}. MS-SSIM is a widely used metric to measure the similarity of paired images, where the higher MS-SSIM the better performance. SWD is an efficient metric to compute random approximation to earth mover's distance, which has also been used for measuring GAN performance, where the lower SWD the better performance. FID calculates the distance between real and fake images at feature level, where the lower FID the better performance. 
 
\subsubsection{Experimental Results:}
\label{Implementation}

\begin{figure}[!t]
	\center
	\includegraphics[width =1\textwidth ]{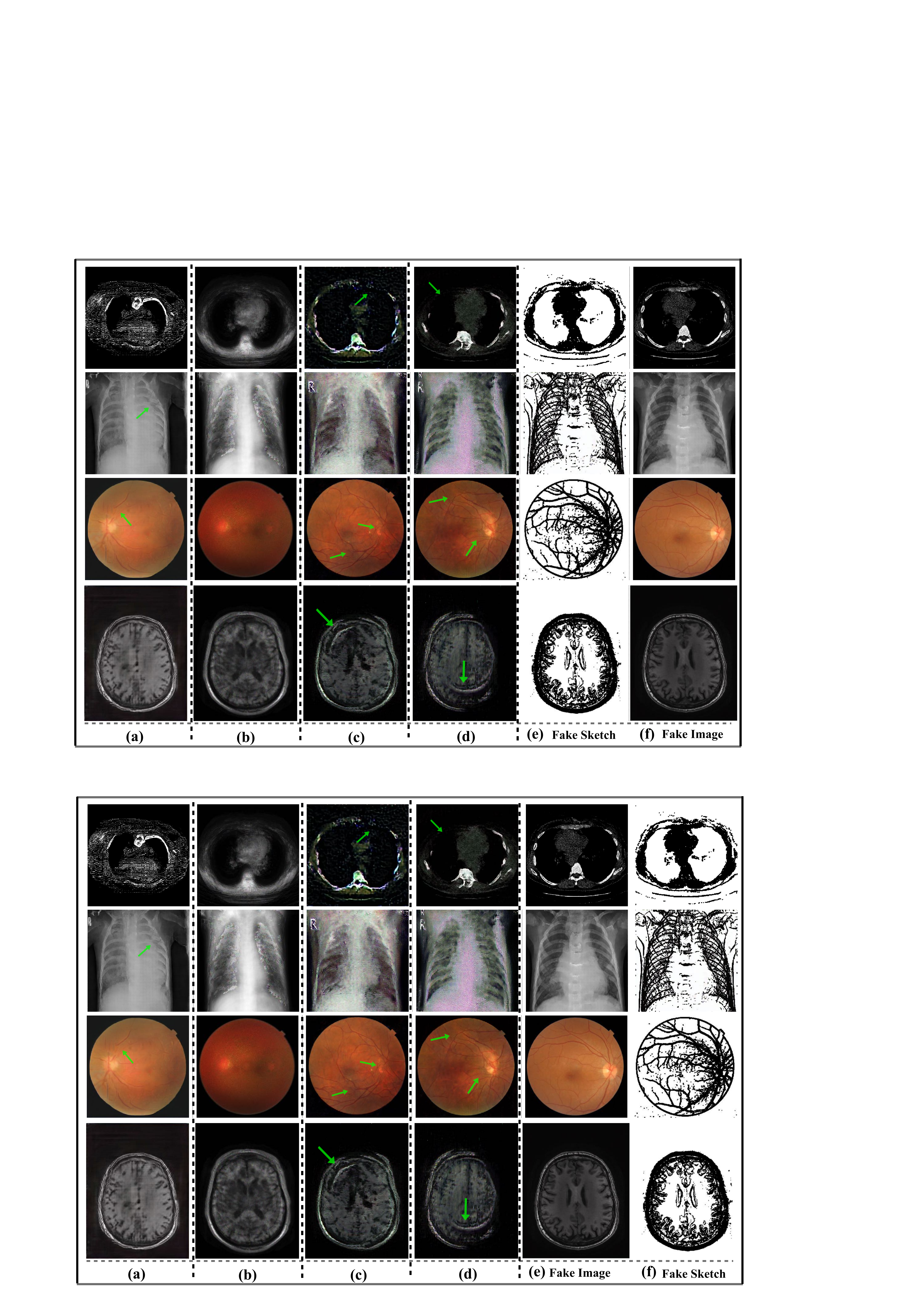} 
	\caption{These images are generated by different GANs: from left to right are results by: (a) PGGAN \cite{PGGAN} , (b) WGAN \cite{WGAN} , (c) DCGAN \cite{DCGAN} , (d) ACGAN \cite{odena2017conditional} and (e) Our SkrGAN . The synthetic sketches generated from random noise are shown in the figure (f). From top to bottom, we show results from:  CT, X-ray, Retina color fundus and MRI. The green arrows illustrate the structural distortions in the generated images. (More visualization results could be found in Supplementary Material.)}
	\label{fig3}
\end{figure}

The images from all datasets are firstly resized to $512\times512\times3$. In $G_S$, $D_S$, $G_P$ and $D_P$, we use Adam optimizers, where the learning rate of $G_S$ and $D_S$ are set to $0.001$, and the learning rate of our $G_P$ and $D_P$ are set to $0.0002$. Based on experience, we set the value of $\lambda$ in Eq (\ref{eq4}) to $100$ and a small change of $\lambda$ does not affect much the performance.  The batch size of our model is set to $16$. The proposed SkrGAN is implemented on PyTorch library with two NVIDIA GPUs (GeForce TITAN XP).


To justify the performance of the proposed method, we compare our SkrGAN with four state-of-the-art GANs:  DCGAN \cite{DCGAN} , ACGAN \cite{odena2017conditional} , WGAN  \cite{WGAN} and PGGAN \cite{PGGAN}. These different methods are used to generate 100 images, and the aforementioned metrics are used  for quantitative comparsions by using these generated images. Table 1 summarizes the results. It can be seen that our SkrGAN achieves SWD of $0.025,~0.026,~0.020$ and $0.028$, MS-SSIM of $0.614$, $0.506$, $0.359$ and $0.436$ and FID of $27.59$, $114.6$, $79.97$ and $27.51$ on the generated retinal color fundus, Chest X-ray, lung CT and brain MRI images, better than other GANs. On one hand, as DCGAN, ACGAN, WGAN and PGGAN are not designed for generating high resolution images from a small dataset. Therefore, these methods produce relatively poor results on generating medical images from small training datasets. On the other hand, these methods only consider the global contextual information and ignore the foreground structures, which lead to the discontinued and distorted sketch structures, such as the discontinued vessel and distorted disc cup in retinal color fundus, the discontinued bones and the distorted lung in chest X-ray, the discontinued ribs in CT and the distorted textures in MRI. By contrast, our method uses sketch to guide the intermediate training step, which guarantees the network to generate high quality medical images with realistic anatomical structures.

Fig. \ref{fig3} illustrates examples of the synthetic images by DCGAN, ACGAN, WGAN, PGGAN, and our method in the four different medical image modalities: CT, X-Ray, retinal color fundus and MRI. It can be observed that SkrGAN presents visually appealing results, where most of the structural features such as the vessel in color fundus, bones in X-ray, ribs and backbone in CT, texture distribution in MRI are close to those in real images. On the contrary, there are some structural distortions in images, which are generated by other GANs, as illustrated by green arrows in Fig \ref{fig3}.

\begin{table}[!t]
\renewcommand\arraystretch{1.2} 
	\scriptsize
	\setlength{\tabcolsep}{1.6 mm}
	\caption{Performances (mean) of different GANs on Retinal color fundus, chest X-Ray, lung CT and Brain MRI. }
	
	\begin{center}
		\begin{tabular}{|l |l |c| c |c |c| c|}
			\hline
			\multicolumn{2}{|c|}{\textbf{Evaluation} }&\multicolumn{5}{|c|}{\textbf{Method}}
			\\ \hline
			Dataset                          & Metric  & SkrGAN     & DCGAN\cite{DCGAN} & ACGAN\cite{odena2017conditional} & WGAN\cite{WGAN} & PGGAN\cite{PGGAN}  \\ 
			\toprule[1pt]
			\multirow{4}{*}{\textbf{\shortstack{Color \\Fundus}}} & SWD $\downarrow$    & \textbf{0.025} & 0.160                               & 0.149                            & 0.078                              & 0.036                          \\ \cline{2-7}
			                                 & MS-SSIM $\uparrow$ & \textbf{0.614} & 0.418                               & 0.490                            & 0.584                              & 0.537 \\ \cline{2-7}
			                                 & FID $\downarrow$     & \textbf{27.59} & 64.83                               & 96.72                            & 240.7                              & 110.8 \\ \hline
			\multirow{4}{*}{\textbf{\shortstack{Chest\\ X-ray}}}   & SWD $\downarrow$    & \textbf{0.026} & 0.118                               & 0.139                            & 0.196                              & 0.031                             \\ \cline{2-7}
			                                 & MS-SSIM $\uparrow$ & \textbf{0.506} & 0.269                               & 0.301                            & 0.401                              & 0.493                             \\ \cline{2-7}
			                                 & FID $\downarrow$    & \textbf{114.6} & 260.3                               & 235.2                            & 300.7                              & 124.2                            \\ \hline
			\multirow{4}{*}{\textbf{\shortstack{Lung \\CT}}}     & SWD $\downarrow$    & \textbf{0.020} & 0.333                               & 0.317                            & 0.236                              & 0.057                              \\ \cline{2-7}
			                                 & MS-SSIM $\uparrow$ & \textbf{0.359} & 0.199                               & 0.235                            & 0.277                              & 0.328 \\ \cline{2-7}
			                                 & FID $\downarrow$    & \textbf{79.97} & 285.0                               & 222.5                            & 349.1                              & 91.89 \\ \hline
			\multirow{4}{*}{\textbf{\shortstack{Brain \\MRI}}}    & SWD $\downarrow$    &            
			\textbf{0.028}
			& 0.163                               & 0.122                            & 0.036                              & 0.042                              \\ \cline{2-7}
			                                 & MS-SSIM $\uparrow$ & \textbf{0.436}           & 0.277                               & 0.235                            & 0.314                              & 0.411 \\ \cline{2-7}
			                                 & FID $\downarrow$    & \textbf{27.51}           & 285.0                               & 222.5                            & 176.1                              & 33.76 \\ \hline
		\end{tabular}
	\end{center}
	\label{table:results}
\end{table}

\subsubsection{Application to Vessel Segmentation:}

Besides the above quantitative and qualitative comparisons, we further apply the proposed SkrGAN as a data augmentation method on a vessel segmentation task in DRIVE\footnote{\url{https://www.isi.uu.nl/Research/Databases/DRIVE/}} \cite{staal:2004-855}  (including 20 training images and 20 testing images). The DRIVE dataset provides two expert manual annotations, and the first one is chosen as the ground truth for performance evaluation in the literature. We generated 2000 synthetic images \begin{wraptable}{r}{8cm}
	\vspace{-40pt}
	\scriptsize
	\center
	\setlength{\tabcolsep}{3.5 mm}
	\caption{ Segmentation performance of U-net}
	\begin{tabular}{|l|c|c|c|}
	\hline
 	{Pretrain}&  SEN     &    ACC   &  AUC       \\ \hline
 	with &  \textbf{ 0.8464}   &  \textbf{0.9513}   &   \textbf{ 0.9762 }   
 	  \\ \cline{1-4}
 	whithout & 0.7781       &  0.9477     &    0.9705  \\ \hline
\end{tabular}
\vspace{-20pt}
\end{wraptable} and utilized the generated sketches as the label to pretrain a vessel detection network. In this paper, we use the U-net \cite{ronneberger2015unet}, which is widely used in many biomedical segmentation tasks. The pretrained model is then further finetuned for vessel detection using 20 training images and tested in 20 testing images. 

To justify the benefits of the synthetic images for training the segmentation network, we compared the trained model using synthetic images with the model without pretraining. The following metrics were calculated to provide an objective evaluation: ${\textit{{sensitivity (SEN)}}}=TP/(TP+FN)$,  ${\textit{{accuracy (ACC)}}} = (TP+TN)/(TP+FP+TN+FN)$, and the Area Under the ROC Curve (\emph{AUC}). The results summarized in Table 2 shows that: pretraining with synthetic images improves \emph{SEN} of the vessel detection by $8.78\%$, while $ACC$ and $AUC$ are improved by pretraining with the synthetic pairs too.

\section{Conclusion}
In this paper, we have proposed an unconditional GAN named Sketching-rendering Unconditional Generative Adversarial Network (SkrGAN) that is capable of generating high quality medical images. Our SkrGAN embedded the sketch representation to guide the unconditional medical image synthesis and generate images with realistic foreground structures. The experiments on four types of medical images, including retinal color fundus, chest X-ray, lung CT and brain MRI, showed that our SkrGAN obtained state-of-the-art performances in medical image synthesis. It demonstrated that the sketch information can benefit the structure generation. Besides, the application of retina vessel segmentation showed that the SkrGAN could be used as a data augmentation method to improve deep network training.

%
%
\bibliographystyle{splncs04}
\bibliography{paper_183}

\end{document}